\documentclass[USenglish]{article}

\usepackage[utf8]{inputenc}
\usepackage[small]{dgruyter}
\usepackage{microtype}
\usepackage{lmodern}
\usepackage[dvipsnames]{xcolor}

\usepackage{float}
\usepackage[T1]{fontenc}
\usepackage{amssymb}
\usepackage{subfig}
\numberwithin{equation}{section}

\usepackage{aliascnt}

\usepackage[backend=biber,maxnames=10,backref=true,hyperref=true,giveninits=true,safeinputenc]{biblatex}
\usepackage{algorithm}
\usepackage{algpseudocode}
\algdef{SE}
[CLASS]
{Class}
{EndClass}
[1]
{\textbf{class} \textsc{#1}}
{\textbf{end class}}

\algdef{SE}
[FUNCTION]
{Function}
{EndFunction}
[1]
{\textbf{function} \textsc{#1}}
{\textbf{end function}}

\usepackage{xcolor}
\def\HiLi{\leavevmode\rlap{\hbox to \linewidth{\color{yellow!50}\leaders\hrule height .8\baselineskip depth .5ex\hfill}}}

\newtheorem{lemma}{Lemma}[section]

\newaliascnt{proposition}{lemma}

\aliascntresetthe{proposition}

\newaliascnt{corollary}{lemma}

\aliascntresetthe{corollary}

\newaliascnt{theorem}{lemma}
\newtheorem{theorem}[theorem]{Theorem}
\aliascntresetthe{theorem}

\newaliascnt{definition}{lemma}
\newtheorem{definition}[definition]{Definition}
\aliascntresetthe{definition}

\newaliascnt{example}{lemma}

\aliascntresetthe{example}

\newaliascnt{assumption}{lemma}

\aliascntresetthe{assumption}

\newaliascnt{notation}{lemma}

\aliascntresetthe{notation}

\newtheorem{remark}{Remark}

\newcommand{\N}{\mathbb{N}}

\newcommand{\R}{\mathbb{R}}

\let\RE\Re
\let\Re=\undefined
\DeclareMathOperator{\Re}{\RE e}
\let\IM\Im
\let\Im=\undefined
\DeclareMathOperator{\Im}{\IM m}

\newcommand{\norm}[1]{\left\|#1\right\|}
\newcommand{\set}[1]{\left\{ #1\right\}}

\newcommand{\bP}{{\bf P}}

\newcommand{\bw}{{\bf w}}

\newcommand{\vx}{{\vec{x}}}
\newcommand{\vy}{{\vec{y}}}

\newcommand{\vp}{{\vec p}}

\begin{document}
	
	
	\author[1]{Leon Frischauf}
	\author[1,2,3]{Otmar Scherzer}
	\author[1,2,*]{Cong Shi}
	\affil[1]{Faculty of Mathematics, University of Vienna, Oskar-Morgenstern-Platz 1, A-1090 Vienna, Austria}
	\affil[2]{Johann Radon Institute for Computational and Applied Mathematics (RICAM), Altenbergerstraße 69, A-4040 Linz, Austria}
	\affil[3]{Christian Doppler Laboratory for Mathematical Modeling and Simulation of Next Generations of Ultrasound Devices (MaMSi), Oskar-Morgenstern-Platz 1, A-1090 Vienna, Austria}
	\title{Classification with neural networks with quadratic decision functions}
	\runningtitle{Quadratic neural networks}

    \abstract{Neural networks with quadratic decision functions have been introduced as alternatives to standard neural networks with affine linear ones. They are advantageous when the
    	objects or classes to be identified are compact and of basic geometries like circles, ellipses etc.
    	In this paper we investigate the use of such ansatz functions for classification.
    	In particular we test and compare the algorithm on the MNIST dataset for classification of handwritten
    	digits and for classification of subspecies. We also show, that the implementation can be based on the neural network structure in the software {\bf \emph{Tensorflow}} and {\bf \emph{Keras}}, respectively.
    }
	
	
	\maketitle
	
\section{Introduction}\label{sec:introduction}
This paper is motivated from the following inverse problems application scenario: We consider the inverse problem of solving the operator equation
\begin{equation} \label{eq:op}
   F(x)=y.
\end{equation}
We are not particularly interested in the solution $x^\dagger$ but in some properties. That means we need to classify $x^\dagger$. A stimulating press article is the identification of an ancient body (Anna Engl \footnote{\url{https://kurier.at/wissen/wissenschaft/identitaet-der-geheimnisvollen-litzlbergerin-geklaert/400829738}}) from which we made up the following scientific problem: Let us assume that we have a non perfect reconstruction of $x^\dagger$, the ideal solution of \autoref{eq:op}, which is a digit from $0$ to $9$. So we have to classify after reconstruction of some approximation of $x^\dagger$. In other word the application we have in mind is solving an inverse problem by a combination of a classical inverse problems solver and machine learning. We see in \autoref{fig:clustering} numbers cluster on compact domains, which is the basis for the mathematical studies afterwards.

\emph{Classification} deals with the problem of associating objects to predefined classes of objects of similar property. It is an essential component of the areas of machine learning and data analysis. The most important classification algorithms are $k$-means and hierarchical classification (see for instance \cite{Nie16}). Recently, \emph{neural networks} have proven to be successful in a wide range of practical applications such as image classification \cite{KriSutHin17}, handwritten digit recognition \cite{LecJacBot95}, speech recognition \cite{HinDenYuDahMoh12,LecBen95}, to mention but a few. Today the success of machine learning methods is driven by the availability of large amounts of training data, efficient software and drastic improvements in computing power. Furthermore, binary classification has been extensively studied in areas such as medical testing, quality control, and information retrieval. Commonly utilized methods for binary classification include decision trees, random forests, Bayesian networks, among others.

The efficiency of neural networks to approximate functions has been rigorously validated over the last three decades. Specifically, a series of these early works \cite{Cyb89, Fun89, HorStiWhi89, Bar93} on universal approximation theorems demonstrated that a continuous function defined on a bounded domain can be approximated by a sufficiently large two-layer neural network. In \cite{FriSchShi24} we discussed approximation properties of generalized neural networks, in particular neural network functions with decision functions, which are at most \emph{quadratic polynomials}.

In this paper we follow up on the topic of classification with neural networks: In contrast to implementations with standard neural networks with \emph{affine linear decision functions}, we investigate neural networks with \emph{quadratic} ones. These have been introduced in \cite{Mha93}, where it was shown that they allow for better approximation of general functions than neural networks with affine linear decision functions. The main difference, and the advantage for applying it to our selected applications is, that the levels sets of neural networks with quadratic decision functions can be compact. In applications neural network functions of such kind have been used in \cite{TsaTefNikPit19}. \cite{FanXioWan20} proposed neural networks with \emph{radial} decision functions, and proved an universal approximation result for ReLU-activated \emph{deep neural networks} with \emph{quadratic} decision functions. Two particular examples are considered in this paper: Digit recognition (see \autoref{fig:clustering}) and subspecies identification (see \autoref{ss:sc}).

\emph{Outline of the paper:} In \autoref{se:background} we make a formal definition of neural networks with quadratic decision functions. In \autoref{sec:simulations} we explain the implementation of classification algorithms with neural networks with quadratic decision functions and show some results in \autoref{sec:results}. As expected, the algorithms perform much better comparing with other classification algorithms, when the cluster to be identified has compact support. More specifically, we investigate letter recognition from the MNIST database \cite{Lec98}. Secondly we compare the performance of deep neural networks with affine linear decision functions with shallow networks with quadratic decision functions for identifying a subpopulation.

\section{Background on neural network functions} \label{se:background}
The purpose of this paper is to investigate the use of neural network functions with \emph{quadratic} decision functions for \emph{classification}. In order to spot the difference to standard neural network functions with affine linear decision functions, we recall their definition first. All along this paper let $m \geq n \in \N$. Line vectors \footnote{These two different notations of vectors are needed so that later results such as \autoref{th:general} can be formulated clearly. The first notation (where components are denoted by superscripts) is reserved for ``weight'' vectors, while $\vx$ is a point, where functions are evaluated.} in $\R^m$ and $\R^n$ are denoted by
\begin{equation*}
  \bw = (w^{(1)},w^{(2)},\ldots,w^{(m)})^T \text{ and } \vx = (x_1,x_2,\ldots,x_n)^T, \text{ respectively.}
\end{equation*}
\begin{definition}[Neural networks with affine linear decision functions] \label{de:affinenns}
A \emph{shallow} neural network with affine linear decision functions ({\bf ALNN}) is a function (here $m=n$)
\begin{equation}\label{eq:classical_approximation}
  \vx \in \R^n \to {\bf p}(\vx) := \Psi[\vp](\vx) := \sum_{j=1}^{N} \alpha_j\sigma\left(\bw_j^T \vx +\theta_j \right),
\end{equation}
with $\alpha_j, \theta_j \in \R$, $\bw_j \in \R^n$; $\sigma:\R \to \R$ is an activation function, such as the sigmoid, ReLU${}^k$ or Softmax function.
Moreover,
\begin{equation*}
  \vp = [\alpha_1,\ldots,\alpha_N; \bw_1,\ldots,\bw_N; \theta_1,\ldots,\theta_N] \in \R^{n_*} \text{ with } n_* = (n+2)N
\end{equation*}
denotes the according parametrization of ${\bf p}$. In this context the set of {\bf ALNN}s is given by
\begin{equation*}
  \bP := \set{{\bf p} \text{ of the form \autoref{eq:classical_approximation}}: \vp \in \R^{n_*}}.
\end{equation*}
\end{definition}
More recently \emph{deep} affine linear neural network functions ({\bf DNN}s) have become popular:
\begin{definition}[Neural networks with deep affine linear decision functions] \label{de:DNNs}\quad
	 An $(L+1)$-layer network looks as follows: 
\begin{equation}\label{eq:DNN}
\begin{aligned}
\vx \in \R^n \to  {\bf p}(\vx) := 
\Psi[\vp](\vx) := {\bf \alpha}^T \vec{\sigma}_L\left( p_{L} \left( \ldots \left( \sigma_1 \left(p_{1}(\vx) \right) \right)\right) \right).
\end{aligned}
\end{equation}
where $$p_{l}(\vx) = (\bw_{1,l},\bw_{2,l},\ldots,\bw_{N_l,l})^T \vx +(\theta_{1,l},\theta_{2,l},\ldots,\theta_{N_l,l})^T$$
with ${\bf \alpha} \in \R^{N_L}, \theta_{j,l} \in \R$ and $\bw_{j,l} \in \R^{N_{l}}$ for all $l=1,\ldots,L$ and all $j=1,\ldots,N_l$.
Here $N_l$ denotes the number of neurons in $l$-th internal layer and $n=N_1$. $\vec{\sigma}_l:\R^{N_l} \to \R^{N_l}$, $l=1,\ldots,L$ denote activation functions for each layer. 
It is common to use the vector valued function 
$$ \vec{\sigma}_l^{(k)}(\vy) = \sigma(\vy^{(k)}),$$
where $.^{(k)}$ denotes the $k$-th component of the vector $\vy$. Moreover, we denote the parametrizing vector by
\begin{equation} \label{eq:pd}
\vp = [{\bf \alpha}; \bw_{1,1},\ldots,\bw_{N_L,L}; \theta_{1,1},\ldots,\theta_{N_L,L}] \in \R^{n_*}.
\end{equation}
\end{definition}
A shallow linear neural network can also be called a \emph{2-layer network}. Note that the output layer $\vp(\vx)$ is also counted as a layer. Therefore, a general $(L+1)$-layer network has only $L$ \emph{internal} layers.

In this paper we use neural network functions with decision functions, which are higher order polynomials, as summarized in \cite{FriSchShi24}, where many more of such families of functions have been considered:

\begin{definition}[Neural networks with radial quadratic decision functions {\bf (RQNN)}] \label{de:rqnn} This is the family of functions of the form
  	\begin{equation}\label{eq:radial_approximation}
  		\begin{aligned}
  			\vx \to {\bf p}(\vx) = \Psi[\vp](\vx) &:=
  			\sum_{j=1}^{N} \alpha_j \sigma \left(\bw_j^T \vx + \xi_j \norm{\vx}^2 +\theta_j\right) \\
  			& \text{ with } \alpha_j, \theta_j \in \R, \bw_j \in \R^n, \xi_j \in \R.
  		\end{aligned}
  	\end{equation}
  	Moreover,
  	\begin{equation*}
  		\begin{aligned}
  			\vp = [\alpha_1,\ldots,\alpha_N; \bw_1,\ldots,\bw_N; \xi_1,\ldots,\xi_N; \theta_1,\ldots,\theta_N] \in \R^{n_*}
  			\text{ with }
  			n_* = (n+3)N
  		\end{aligned}		
  	\end{equation*}
  	denotes the according parametrization of ${\bf p}$.
  	In this context the set of {\bf RQNN}s is given by
  	\begin{equation*}
  		\bP := \set{{\bf p} \text{ of the form \autoref{eq:radial_approximation}}: \vp \in \R^{n_*}}.
  	\end{equation*}
  	The composition of $\sigma$ with the polynomial is called \emph{neuron}. 
  	
  	{\bf RQNN}s are shallow. Deep variants of it are denoted by {\bf DRQNN}s. The terminology radial is due to the term $\xi_j \norm{\vx}^2$. If in the decision function, instead of $\xi_j \norm{\vx}^2$ a more general function $ \norm{A_j \vx}^2$ with $A_j \in \R^{n \times n}$ would be used, then the term radial would be omitted.
  \end{definition}

In order to show the feasibility of quadratic neurons, we recall two theoretical results:
  	\begin{lemma}[$L^\infty$-convergence \cite{FriSchShi24}]\label{th:general}
  	Let $\sigma:\R \to \R$ be a continuous discriminatory function \footnote{A function $\sigma : \R \to \R$ is called \emph{discriminatory} (see \cite{Cyb89}) if every measure $\mu$ on $[0,1]^n$, which satisfies $\int_{[0,1]^n} \sigma (\bw^T \vx + \theta)\,d\mu(\vx) = 0$ for all $\bw \in \R^n$ and $\theta \in \R$ implies that $\mu \equiv 0$.}.
  	Then for every $f\in C([0,1]^n)$ and every $\epsilon>0$, there exists $N\in \N$ and $\vp \in \R^{n_*}$ 
  	such that
  	\begin{equation*}
  		\norm{f-\Psi[\vp]}_{L^\infty} < \epsilon \text{ for all }\vx\in [0,1]^n,
  	\end{equation*}
    where $\Psi[\vp]$ is a {\bf RQNN} from \autoref{eq:radial_approximation}.
   \end{lemma}

  	\begin{theorem}[$L^2$-convergence rate of {\bf RQNN} \cite{FriSchShi24}] \label{co:ConvR}
  		For every function $f \in \mathcal{L}^1$ (in the sense of \cite{ShaCloCoi18}) and every $N \in \N$, there exists a parametrization vector $\vp$ 
  		such that
  		\begin{equation}\label{eq:app_error II}
  			\norm{f-\Psi[\vp]}_{L^2} = \mathcal{O}( (N+1)^{-1/2}),
  		\end{equation}
  		where $\Psi[\vp]$ is a {\bf RQNN} from \autoref{eq:radial_approximation}.
  \end{theorem}

  \begin{remark}
  	\begin{itemize}
  		\item In \autoref{th:general}, the parameter $N$ is determined by $\epsilon$ and $f$.
  		\item \autoref{co:ConvR} gives the approximation rate under the assumption that $f\in \mathcal{L}^1$, and \autoref{th:general} only provides the convergence.
  		\item In comparison with affine linear neural networks, some constrains induce that the level-sets of the quadratic neurons are compact and not unbounded, as they are for affine linear neurons. 
  	\end{itemize}
  \end{remark}

  \begin{remark}
  	For $\xi_j \neq 0$ the argument in \autoref{eq:radial_approximation} rewrites to
  	\begin{equation}\label{eq:nus}
  		\begin{aligned}
  			\nu_j(\vx) := \xi_j \norm{\vx}^2 + \hat{\bw}_j^T\vx  + \theta_j
  			= \xi_j \norm{\vx - \vy}^2 + \kappa_j
  		\end{aligned}
  	\end{equation}
  	with
  	\begin{equation} \label{eq:kappa}
  		\vy_j = - \frac{1}{2\xi_j} \hat{\bw}_j^T \text{ and } \kappa_j =
  		\theta_j - \frac{\norm{\hat{\bw}_j}^2}{4 \xi_j}.
  	\end{equation}
  	We call the set of functions from \autoref{eq:radial_approximation}, which satisfy
  	\begin{equation} \label{eq:circle}
  		\xi_j \geq 0 \text{ and } \kappa_j \leq 0 \; \text{ for all } j=1,\ldots,N,
  	\end{equation}
  	neural network with \emph{circular decision functions}, because the level sets of $\nu_j$ are circles. These are \emph{radial basis functions}
  	(see for instance \cite{Buh03} for a general overview).
  \end{remark}
  The focus of this paper is to investigate the use of {\bf RQNN}s for solving classification problems. The main contribution is to show that {\bf RQNN}s are compatible in performance with standard {\bf DNN}s, and that the implementation is straight forward in standard machine learning software such as {\bf \emph{Tensorflow}} \cite{AbaAgaBamBreChe16_report} and {\bf \emph{Keras}} \cite{Cho15}. In other words we show that \emph{shallow quadratic} neural networks are comparable with \emph{deep linear} neural networks in some classification applications.

    \section{Binary classification}\label{sec:simulations}

    \emph{Clustering} refers to the task of identifying groups of data with similar properties in a large dataset. It is typically referred to \emph{unsupervised learning}. \emph{Classification} on the other hand uses \emph{labeled data}
    (such as for instance obtained by cluster analysis) and is referred to
    \emph{supervised learning}.

    \subsection{Clustering}
    \begin{definition}[Clustering] \label{de:clust}
    	Given a dataset $P=\set{\vx_1, \vx_2, ...,\vx_N} \in \R^n$ and a target number of clusters $k$, the aim is to find a
    	partition $C_1,\ldots,C_k$ of $\R^n$, that is
    	\begin{equation*}
    		C_i \cap C_j=\emptyset \text{ for all } i \neq j \in \set{1,\ldots,k} \text{ and }
    		\bigcup_{i=1}^k C_i = \R^n,
    	\end{equation*}
    	with \emph{centroids}
    	\begin{equation*}
    		\vx_i^c := \frac{1}{|C_i|}\sum_{\vx_j\in C_i}\vx_j \quad  \text{ for all } i =1,\ldots,k,
    	\end{equation*}
    	such that the functional
    	\begin{equation*}
    		(C_1,\ldots,C_k) \mapsto \sum_{i=1}^k \sum_{\vx_j \in C_i} \norm{\vx_j-\vx_i^c}^2
    	\end{equation*}
    	becomes minimal.
    \end{definition}

    The most prominent clustering algorithms are $k$-means and hierarchical clustering (see for instance \cite{Nie16}).
    \begin{itemize}
    	\item The $k$-means algorithms solves the problem from \autoref{de:clust}. However, this problem is NP-hard, and in practice heuristics, such as the Voronoy cell decomposition (see for instance \cite{Mit97}), have to be implemented.
    	\item Hierarchical clustering methods \cite{Nie16} aim at computing a hierarchy of clusters. There are two distinct approaches: agglomerative or divisive hierarchies. The first is a ``bottom-up'' approach, where each data point starts in its own cluster, and clusters are then merged together at each subsequent step. The second follows a ``top-down'' approach, where all data points start in the same cluster, and at each step the clusters are refined with subdivisions.
    \end{itemize}
    While these methods work rather well for diverse practical applications, they do not always provide optimal or even stable solutions. For instance, the $k$-means approach is very sensitive to the choices of initial centroids.

    \subsection{Classification}
    Clustering does not assume knowledge of some elements of $P$ to be labeled. That is, we do not assume the
    knowledge of the labels $i=1,\ldots,k$ associated to some category of elements in $P$. 
    \begin{definition}[Classification] \label{de:class} Let $N_0, N \in \N$ with $N_0 \leq N$ and a dataset $P=\set{\vx_1, \vx_2, ...,\vx_N} \subseteq \R^n$.
    \begin{itemize}
    	\item We are giving the \emph{training data} $P_0 = \set{(\vx_j,l_j):j=1,\ldots,N_0}$, which consists of pairs of elements of $P$ and associated labels $l_j \in \set{1,\ldots,k}$, describing a category. We assume that $\vx_j$, $j=1,\ldots,N_0$ are pairwise different. 
    	\item $P \backslash P_0$ is called \emph{test data}.
    \end{itemize}
    The goal is to find a partition $C_1,\ldots,C_k$ of $\R^n$, such that
    \begin{equation*}
        \vx_j \in C_s \text{ if } l_j =s \text{ for all } j=1,\ldots,N \text{ and } s=1,\ldots,k.
    \end{equation*}
    Finally each element of $P$ can be assigned a label according to the set it is an element of.
    \end{definition}

    \subsection*{Classification with neural networks}
   
    The general idea behind classification is to construct a neural network function $f$ from the set of data points (that is from $\R^n$) to the set of labels $\set{1,\ldots,k}$
    \begin{equation} \label{eq:nnk}
    	f : \R^n \to \set{1,\ldots,k},
    \end{equation}
    where the coefficients of the neurons are optimized using training data $P_0$. Afterwards for a given arbitrary point $\vx$ the label $f(\vx)$ is associated to $\vx$. In our setting, $f$ is a composition of a neural network structure and a decision function (we mostly use threshold functions in contrast to the decision functions used for defining neural networks). The specified layer structure of the network can be an {\bf ALNN} as defined in \autoref{eq:classical_approximation} or an {\bf RQNN} as defined in \autoref{eq:radial_approximation}. $f$ is visualized in \autoref{shallow_netw} and \autoref{deep_netw}.

    In the training process, the output can be seen as a probability, determining whether a point belongs to a specific class or not. For instance, in binary classification for numbers, one class may denote whether the point corresponds to a digit 4, while the other class indicates it does not.

    Of course also deep neural network structures can be used, such as {\bf DNN}s or {\bf DRQNN}s, but this is not the main topic of this paper.

    \section{Binary classification}\label{sec:results}
    In the following we study two binary classification experiments with {\bf RQNN}s.
    In the two examples the data points are in $\R^2$ (see \autoref{origdata} and \autoref{fig:clustering}).
    In supervised learning, the training data is labeled. The goal is to assign each element a category
    (``Species 1'' and ``Species 2'').
    There are numerous machine learning techniques available to tackle binary classification issues, from k-means to advanced deep learning models. The optimal choice for performance, considering both speed and accuracy, varies depending on factors like the size and complexity of the data (such as the number of samples and features) and the quality of the data (considering aspects like outliers and imbalanced datasets).

    By induction one may extend binary classification to classification with more than two species by simply rerunning the binary classification algorithms multiple times.

    With neural networks the concrete problem of binary classification of 2D image data is to find a function
    $f$ as in \autoref{eq:nnk} with $k=2$ and $n=2$. Thus
    \begin{equation*}
    	\begin{aligned}
    	 f: \R^2 &\to \set{1,2} \\
    	 (x_1, x_2) = \vx &\mapsto l
    	\end{aligned}
    \end{equation*}
    where $x_1$ and $x_2$ are two coordinates (this are the two input neurons) of one data point in $\R^{2}$ to its label $l$
    (this is the output neuron); see \autoref{shallow_netw} and \autoref{deep_netw}.
    In our test the function $f$ is an {\bf ALNN} (see \autoref{eq:classical_approximation}), {\bf DNN} (see \autoref{eq:DNN}),
    {\bf RQNN} (see \autoref{eq:radial_approximation}) or {\bf DRQNN} (see \autoref{de:rqnn}). The function $f$ is determined in
    such a way that it optimally represents the input-output relation of training data and according labels (see \emph{Algorithm 1}).

    \subsection{Implementation of {\bf RQNN}s}
    The implementation of the binary classification algorithm with neural networks with radial quadratic decision functions ({\bf RQNN}s) can be implemented with \textbf{TensorFlow} \cite{AbaAgaBamBreChe16_report} as well as \textbf{Keras} \cite{Cho15} by \emph{customized layers} in a straight forward manner.

    In the following we outline the pseudocode for concrete implementations for training data in \emph{Algorithm 1} and for test data in \emph{Algorithm 2:}
    \begin{algorithm}
    	\caption*{\emph{Algorithm 1:} The binary classification algorithm with {\bf RQNNs} for training}\label{alg:training}
    	\begin{algorithmic}
    		
    		\State \textbf{Inputs:}
    		\State $\quad$ $\vx$ 
    		coordinates of the training data points.
    		\State $\quad$ input\_dimension $(x_1, x_2)$ $\leftarrow$ number of neurons of $\vx$
    		\State $\quad$ output\_dimension $l$ $\leftarrow$ number of neurons of output
    		
    		\State \textbf{Variables:}
    		
    		\State  $\quad$  $\bw_j$ \ $ \leftarrow$ (trainable) weights for the ``affine linear'' part
    		\State $\quad$  \HiLi $\xi_j$ $\leftarrow$ (trainable) weights for the ``circular'' part
    		\State  $\quad$  $\theta_j$ $\leftarrow$ bias for each neuron
    		
    		\State \textbf{Initialize:}
    		\State  $\quad$  $\bw_j$, $\xi_j$  $\leftarrow$ random Gaussian distribution
    		\State  $\quad$  $\theta_j$ $\leftarrow$ 0
    		
    		\Class{Circular\_layer(input\_dimension, output\_dimension)}
    		
    		\Function{call(inputs)}
    		\State \HiLi outputs $\leftarrow$ $\bw_j^T \cdot \vx + \xi_j \norm{\vx}^2 +\theta_j$
    		
    		\State     store outputs
    		\State    return outputs
    		\EndFunction
    		
    		\Function{backprop(labels):}
    		\State update weights with Adam optimizer with a learning rate of $0.001$,  the number of iterations of the Adam optimizer corresponds to epochs number.
    		\EndFunction
    		\EndClass
    	\end{algorithmic}
    \end{algorithm}

    \begin{algorithm}
	\caption*{\emph{Algorithm 2:} The binary classification algorithm with {\bf RQNNs} for test data}
	\begin{algorithmic}
		
		\State \textbf{Inputs:}
		\State $\quad$ $\vx$ 
		coordinates of the test data points.

		\State \textbf{Parameters:} Trained weights from training process.
					
		\Function{call(inputs)}
		\State \HiLi outputs $\leftarrow$ model.predict($\vx$) (apply trained model $\bw_j^T \cdot \vx + \xi_j \norm{\vx}^2 +\theta_j$ onto test data)
		
		\While {$i \in \text{outputs}$ }

		 \If{$i>0.5$}
  		 \State Point belongs to ``Species 1''

  		 \Else
    		 \State Point belongs to ``Species 2''

  		 \EndIf
		 \State choose next $i$
		 		 \EndWhile

		\EndFunction
		
	\end{algorithmic}\label{alg:test}
\end{algorithm}

    The implementation of {\bf RQNN}s deviates from the implementation with a standard shallow affine linear neural network {\bf ALNN} at the marked positions, where the code needs to be customized.

    \subsection{Subspecies classification} \label{ss:sc}
\begin{figure}[H]
    	\centering
    	\begin{minipage}[b]{0.7\textwidth}
    		\includegraphics[width=\textwidth]{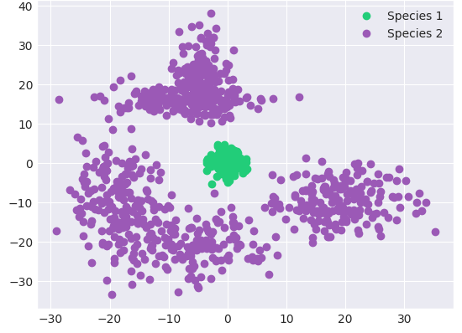}
    		\caption{Ground truth data for the subspecies classification experiment. Only part of the $5000$ training data are shown.}\label{origdata}
    	\end{minipage}
    	\hfill
    \end{figure}

    The goal of the following experiment is to classify a sub-species in a larger population:
    We randomly computed $6000$ points in $\R^2$. Each point was assigned to a species (denote by purple and green as shown in \autoref{origdata}). Each point has a label: subspecies is green and main species is purple (see \autoref{origdata}).

\begin{figure}
    	\centering
    	\begin{minipage}[b]{0.48\textwidth}
    		\includegraphics[width=\textwidth]{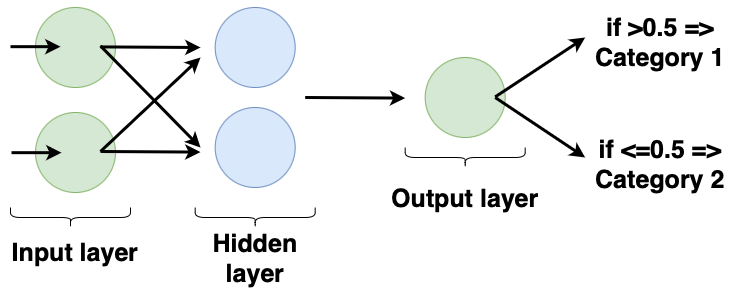}
    		\caption{Neural network structure of a shallow neural network. The output layer gives a number, can be interpreted as a probability, on which classification is performed by taking a threshold. $f$ in \autoref{eq:nnk} maps input to category label, composing a neural and a decision function.} \label{shallow_netw}
    	\end{minipage}
    	\hfill
    	\begin{minipage}[b]{0.48\textwidth}
    		\includegraphics[width=\textwidth]{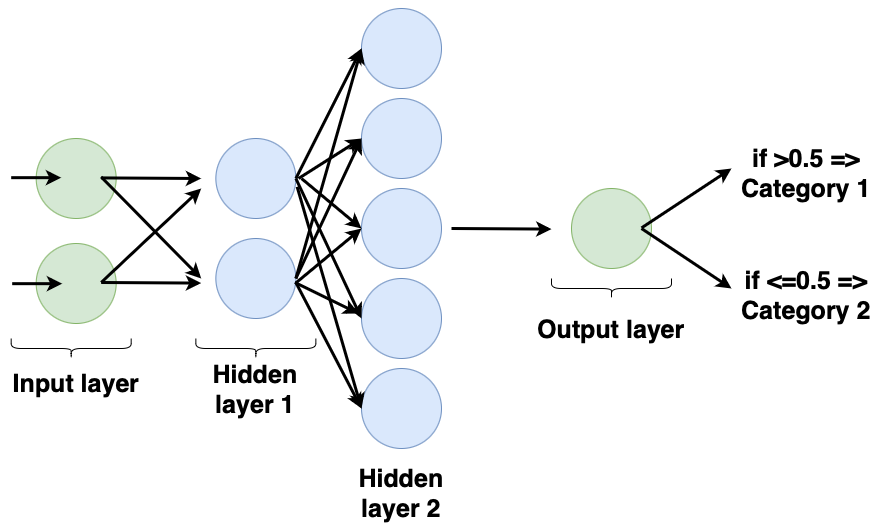}
    		\caption{Neural network structure of a deep neural network. } \label{deep_netw}
    	\end{minipage}
    \end{figure}

    We have split the $6000$ data points into $5000$ training data and $1000$ test data points. The selection of which data point corresponds to the training or test data has been done randomly.
    We experiment with different layer structures of neural networks for binary classification; see \autoref{shallow_netw} for visualization of a shallow neural network and \autoref{deep_netw} for a deep neural network, respectively. Additionally, we have conducted experiments using
    the k-means classification algorithm to compare effectiveness.
    The output layer contains a single neuron; classification is based on thresholding the values of the output neuron: If the number is larger than $0.5$ it belongs to ``Species 1'', otherwise to ``Species 2''.

\begin{figure}[H]
    	\centering
    	\begin{minipage}[b]{0.48\textwidth}
    		\includegraphics[width=\textwidth]{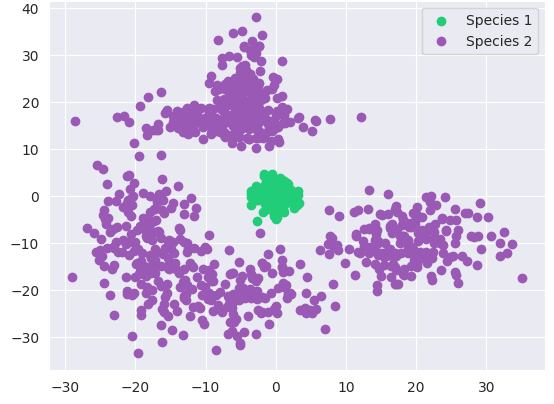}
    		\caption{Classification via {\bf RQNN} and $10$ training epochs} \label{class1}
    	\end{minipage}
    	\hfill
    	\begin{minipage}[b]{0.48\textwidth}
    		\includegraphics[width=\textwidth]{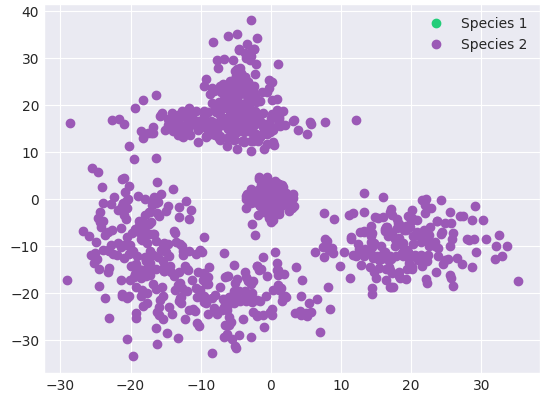}
    		\caption{Classification via {\bf ALNN} and $10$ training epochs} \label{class2}
    	\end{minipage}
    	\hfill
    	\begin{minipage}[b]{0.48\textwidth}
    		\includegraphics[width=\textwidth]{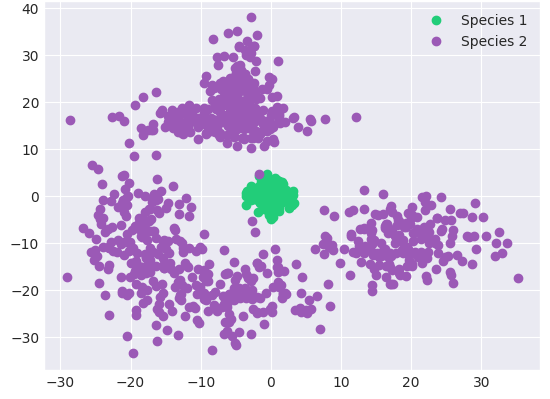}
    		\caption{Classification via two hidden layers of {\bf DNN} and $10$ training epochs} \label{class3}
    	\end{minipage}
    	\hfill
    	\begin{minipage}[b]{0.48\textwidth}
    		\includegraphics[width=\textwidth]{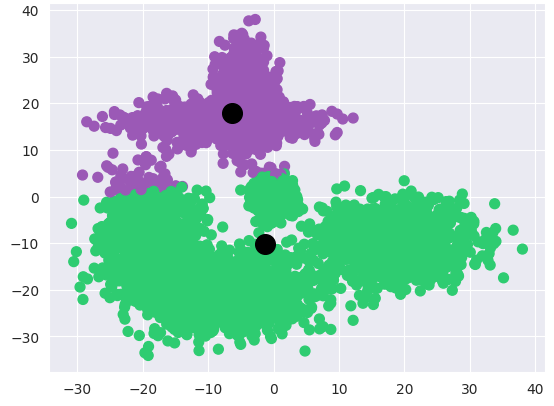}
    		\caption{Classification with k-means with 2 clusters}  \label{class4}
    	\end{minipage}
    \end{figure}

\begin{figure}[H]
    	\centering
    	\begin{minipage}[b]{0.7\textwidth}
    		\includegraphics[width=\textwidth]{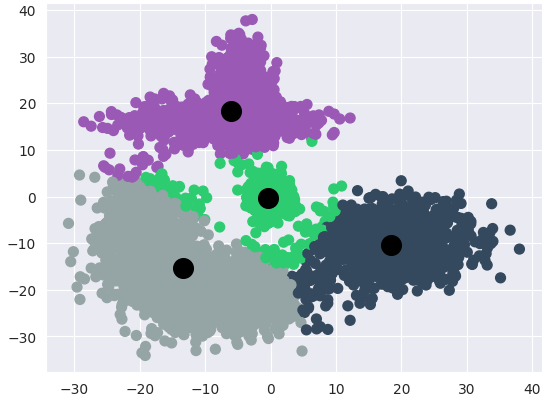}
    		\caption{Classification with k-means with 4 clusters}  \label{class5}
    	\end{minipage}
    	\hfill
    \end{figure}

    In \autoref{class1} - \autoref{class4} one sees the results of the binary classification \emph{Algorithm 2}. Training was done with \emph{Algorithm 1} with $10$ epochs - one epoch stands for one iteration of the Adam optimization algorithm.  From the second layer to $L$th layer we are using ReLU activation functions and for the $(L+1)$th layer Softmax. The initialization of weights in training \emph{Algorithm 1} is performed with Gaussian random values.
    One observes that the classification via {\bf RQNN} works perfect (see \autoref{class1}), i.e., it reaches an accuracy of $100 \%$. When using shallow {\bf ALNN} it does not lead to reasonable classification (see \autoref{class2}). This could happen as it is prioritizing species 2, which is predominantly used within the example. With only one linear layer, it cannot effectively cluster species 2 without losing too much precision on species 1. When using e.g. more epochs, multi-hidden-layer {\bf DNN}s are competitive with shallow {\bf RQNN} (see \autoref{class3}). The k-means algorithm does not yield effective results because it intends to split $\R^2$ into two non-compact domains. It is only somehow competitive when we increase the number of subspecies to four (see \autoref{class5}). This approach necessitates manual merging of subspecies via post-processing, indicating a limitation in its ability to directly classify the subspecies. However, one advantage of k-means is that it does not require the data labels, because it is an unsupervised method.

    The full results can be found in \autoref{tab:comp}. 
   
    \begin{table}
    	\centering
    	\begin{tabular}{c|c|c|c|c|c}
    		Type
    		&
    		Epochs
		&
		Trainable weights
    		&
        Depth
		&
		Width
    		&
    		Accuracy \\
    		\hline
    		{\bf RQNN} & 10 & 3 & 1 & - & 1 \\
    		{\bf ALNN}  &   10 & 2 & 1 & - & 0.823 \\
    		{\bf DNN}   &   10  & 4 & 2 & 5 & 0.997 \\
    		\hline
    		{\bf k-means} (2 clusters)& - & - & - & - & 0.508 \\
    		{\bf k-means} (4 clusters)& - & - & - & - & 0.815 \\
    	\end{tabular}
    	\caption{\label{tab:comp} Overview of the result of the numerical experiments. It indicates the advantage of using {\bf RQNN}s in comparison with {\bf ALNN}s for this scenario. One has to use {\bf DNN}s to match the performance of an {\bf RQNN}. The accuracy is calculated by the percentage of points classified correctly.}
    \end{table}

\begin{figure}[H]
    	\centering
    	\begin{minipage}[b]{0.48\textwidth}
    		\includegraphics[width=\textwidth]{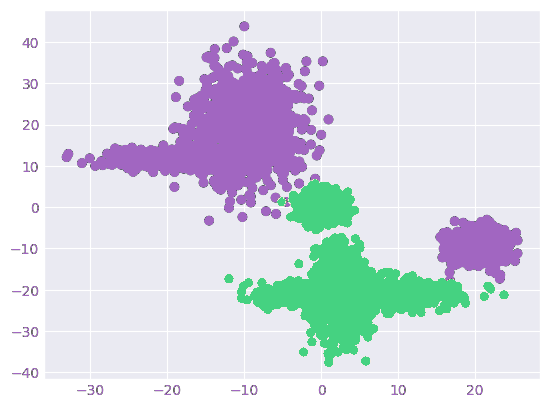}
    		\caption{Truth data of example with no circular shape} \label{not_circular_1}
    	\end{minipage}
    	\hfill
    	\begin{minipage}[b]{0.48\textwidth}
    		\includegraphics[width=\textwidth]{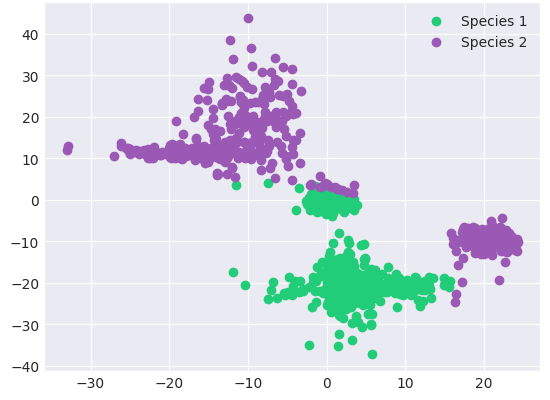}
    		\caption{Classification via {\bf RQNN} and $10$ training epochs (Accuracy: 0.986)} \label{not_circular_2}
    	\end{minipage}
    	\hfill
    	\begin{minipage}[b]{0.48\textwidth}
    		\includegraphics[width=\textwidth]{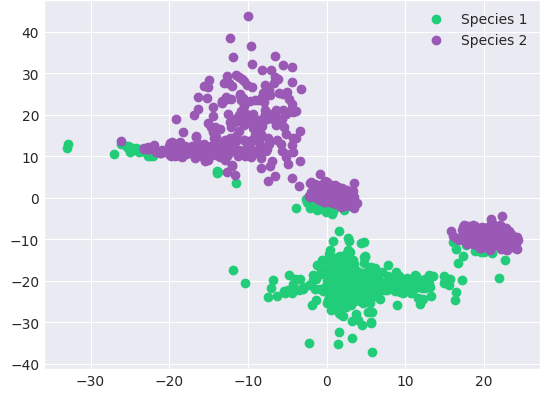}
		\caption{Classification via {\bf ALNN} and $10$ training epochs \\(Accuracy: 0.841)} \label{not_circular_3}
    	\end{minipage}
    	\hfill
    	\begin{minipage}[b]{0.48\textwidth}
    		\includegraphics[width=\textwidth]{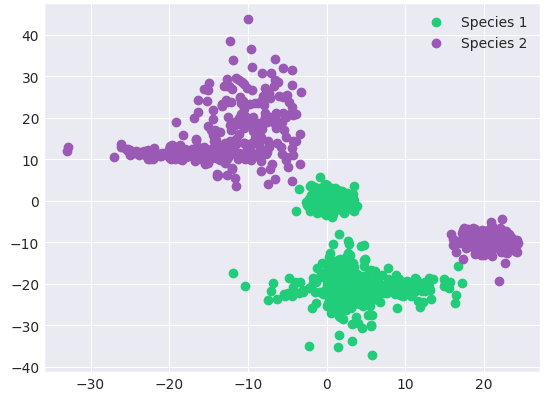}
    		\caption{Classification via two hidden layers of {\bf DNN} and $10$ training epochs \\(Accuracy: 0.998)} \label{not_circular_4}
    	\end{minipage}
    \end{figure}

Finally, we want to have a look at one example (see \autoref{not_circular_1}), where the class does not have a circular shape. The parameters are kept the same as in the previous experiment. The classification with a circular layer (see \autoref{not_circular_2}) works considerably better than with a linear layer (see \autoref{not_circular_3}). This could be due to the higher flexibility of the circular layer compared to the linear one. Only when using a {\bf DNN} one gets a higher accuracy (see \autoref{not_circular_4}).

    \subsection{Binary classification of handwritten digits}

    For a more practical example, we consider the MNIST dataset (see \cite{Lec98}) consisting of 70.000 handwritten digits. We have used $60 000$ data points for training the neural network and $10 000$ data points for testing.

\begin{figure}[H]
    	\centering
    	\includegraphics[width=0.8\textwidth]{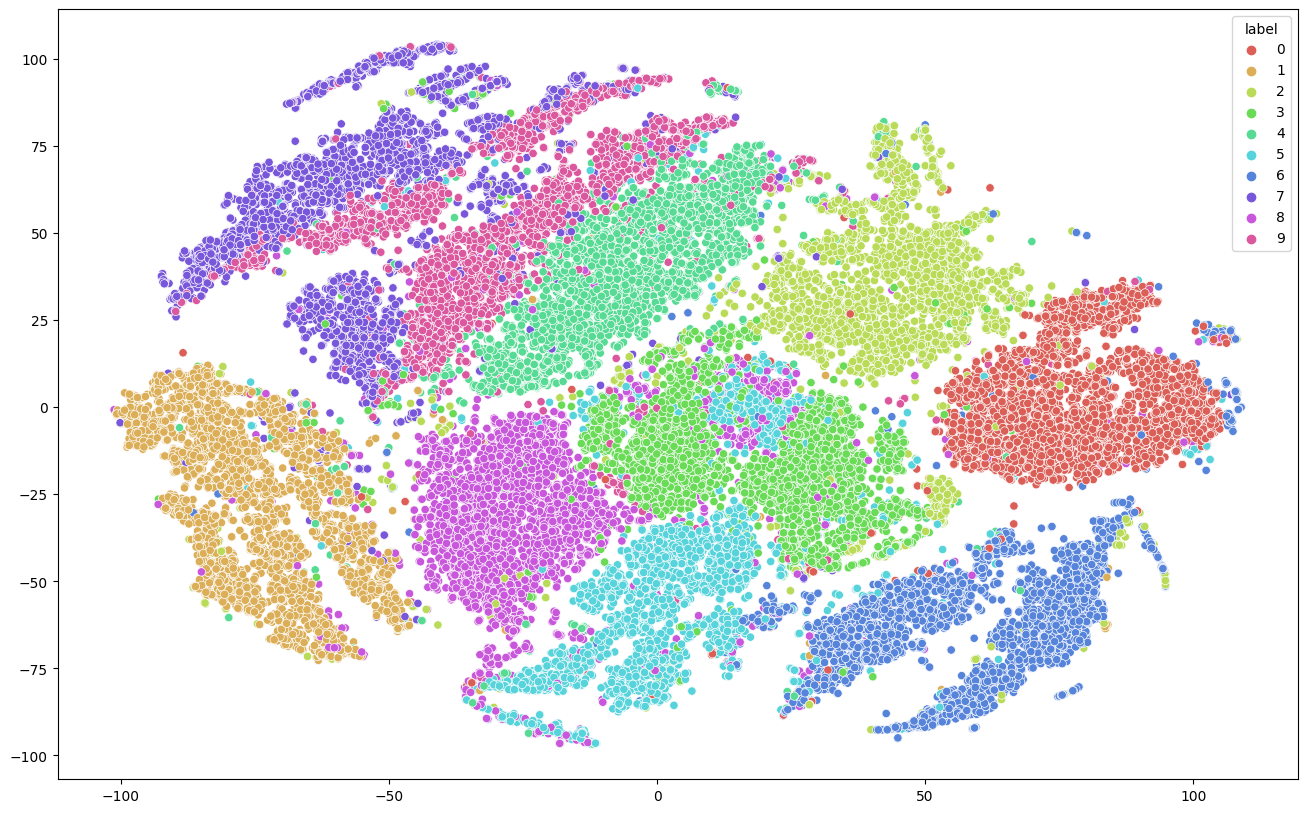}
    	\caption{\label{fig:clustering} (t-SNE) embedding of MNIST shows visualization of numbers in 2 dimension. Different colors correspond to different numbers ($0-9$).}	
    \end{figure}

    The dataset is visualized in $\R^2$ with the t-distributed stochastic neighbor embedding (t-SNE) (see \cite{MaaHin08})\footnote{\url{https://towardsdatascience.com/dimensionality-reduction-using-t-distributed-stochastic-neighbor-embedding-t-sne-on-the-mnist-9d36a3dd4521}.}. In \autoref{fig:clustering} one sees the visualization of MNIST data in $\R^2$, which has been used for
    training in \emph{Algorithm 1}.

\begin{figure}[H]
    	\centering
    	\begin{minipage}[b]{0.48\textwidth}
    		\includegraphics[width=\textwidth]{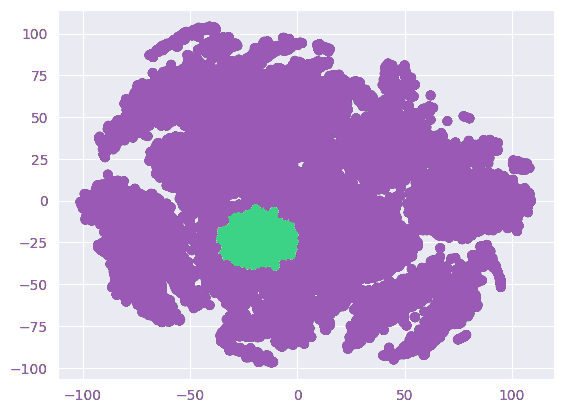}
    		\caption{Binary-classification via {\bf RQNN} and $10$ training epochs. The green area corresponds to the digit 8, the purple area of not being digit 8.}\label{shallow_ell}
    	\end{minipage}
    	\hfill
    	\begin{minipage}[b]{0.48\textwidth}
    		\includegraphics[width=\textwidth]{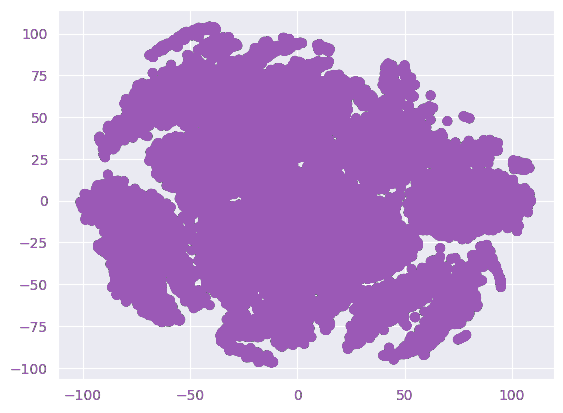}
    		\caption{Binary-classification via {\bf ALNN} and $10$ training epochs. With only one linear layer, it could not cluster the digit $8$, so there is no green area.}\label{shallow_lin}
    	\end{minipage}
    	\hfill
    	\begin{minipage}[b]{0.48\textwidth}
    		\includegraphics[width=\textwidth]{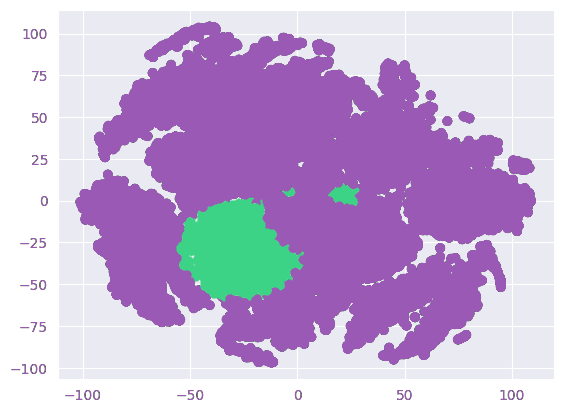}
    		\caption{Binary-classification via {\bf DRQNN} and $30$ training epochs. The green area corresponds to the digit 8, the purple area of not being digit 8.}\label{deep_ell}
    	\end{minipage}
    	\hfill
    	\begin{minipage}[b]{0.48\textwidth}
    		\includegraphics[width=\textwidth]{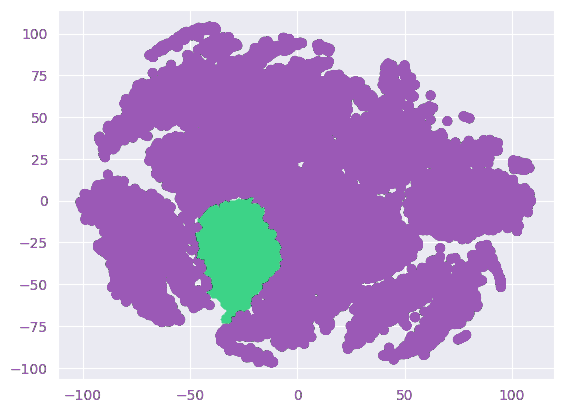}
    		\caption{Binary-classification via {\bf DNN} and $30$ training epochs. The green area corresponds to the digit 8, the purple area of not being digit 8.}\label{deep_lin}
    	\end{minipage}

    \end{figure}

\begin{figure}[H]
    	\centering
    	\begin{minipage}[b]{0.6\textwidth}
    		\includegraphics[width=\textwidth]{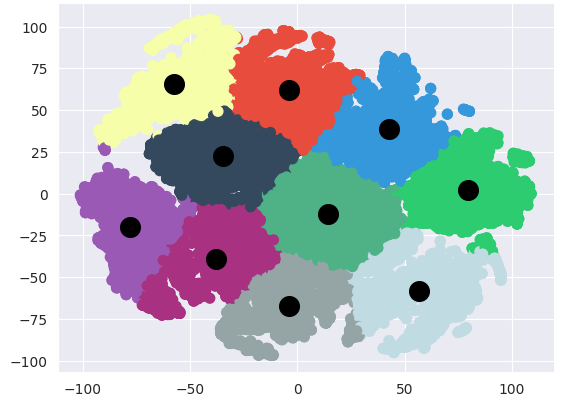}
    		    		\caption{Classification with k-means with 10 clusters. Each digit represents a cluster. In order to get a reasonable cluster for digit $8$ we have to overestimate the number $k$ of clusters.}
\label{kmeans_mnist}
    	\end{minipage}
    	\hfill

    \end{figure}

    The goal is to classify the number 8 versus the others, which is a binary classification problem.
Different neural network structures, for example, {\bf RQNN}, {\bf ALNN}, and {\bf DNN} have been compared with k-means, where the results are illustrated in \autoref{tab:comp_deep}. It indicates the advantage of using {\bf DRQNN}s (see \autoref{deep_ell}) or {\bf RQNN}s (see \autoref{shallow_ell}) in comparison with {\bf ALNN}s (see \autoref{shallow_lin}). {\bf ALNN}s do not lead to sufficiently good results, so one has to use at least {\bf DNN}s (see \autoref{deep_lin}). In \autoref{kmeans_mnist} one observes the clustering performed with k-means. The full comparison can be seen in \autoref{tab:comp_deep}. The visualizations of the classification are provided in \autoref{deep_netw}.
    \begin{table}
    	\centering
    	\begin{tabular}{c|c|c|c|c|c}
    		Type
    		&
    		Epochs
		&
		Trainable weights
		&
		Depth
    		&
    		Width
    		&
    		Accuracy (Crossentropy)\\
    		\hline
    		{\bf RQNN} & 10 & 3  & 1 & - & 0.9250 \\
    		{\bf DRQNN}  &  10 & 9  & 3 & 5 & 0.9616 \\
    		{\bf DRQNN}   &  30 & 9  & 3 & 20 & 0.9630 \\
    		\hline
    		{\bf ALNN} & 10 & 2 & 1 & - & 0.9023 \\
    		{\bf DNN} &   10  & 6 & 3 & 5 & 0.9593 \\
    		{\bf DNN}  &   30 & 6 & 3 & 20 & 0.9766 \\
    	\end{tabular}
    	\caption{\label{tab:comp_deep} Overview of the result of the numerical experiments for deep networks. It indicates the advantage of using {\bf RQNN}s in comparison with {\bf ALNN}s for this scenario.}
    \end{table}

    Finally, we want to classify the number $7$ to see how well {\bf RQNN}s or {\bf DRQNN}s perform when there is no circular shape.

\begin{figure}[H]
    	\centering
    	\begin{minipage}[b]{0.48\textwidth}
    		\includegraphics[width=\textwidth]{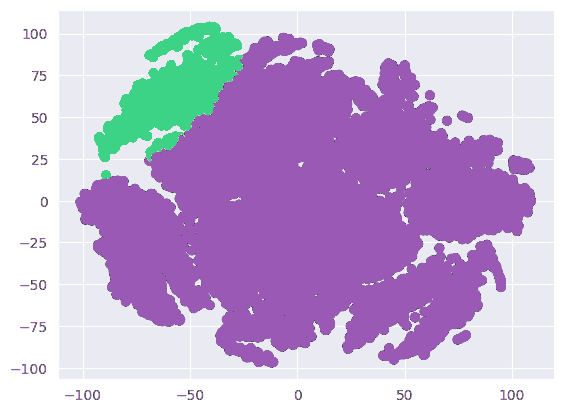}
    		\caption{Binary-classification via {\bf RQNN} and $10$ training epochs. The green area corresponds to the digit $7$, the purple area of not being digit $7$.}\label{not_circular_dig7_1}
    	\end{minipage}
    	\hfill
    	\begin{minipage}[b]{0.48\textwidth}
    		\includegraphics[width=\textwidth]{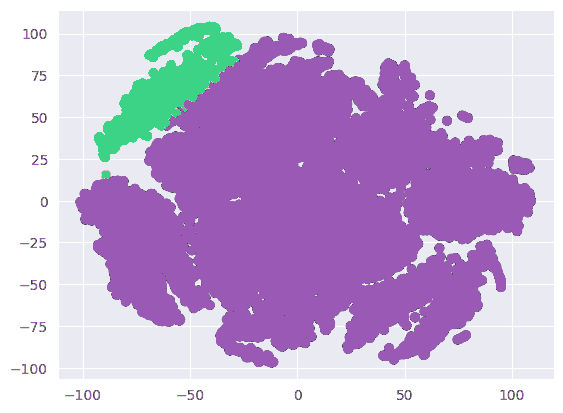}
    		\caption{Binary-classification via {\bf ALNN} and $10$ training epochs. The green area corresponds to the digit $7$, the purple area of not being digit $7$.}\label{not_circular_dig7_2}
    	\end{minipage}
    	\hfill
    	\begin{minipage}[b]{0.48\textwidth}
    		\includegraphics[width=\textwidth]{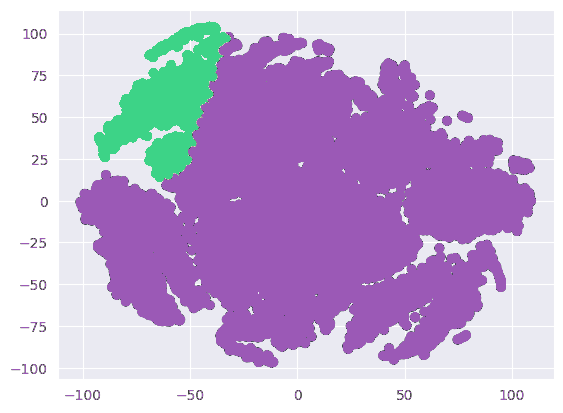}
    		\caption{Binary-classification via {\bf DRQNN} and $10$ training epochs. The green area corresponds to the digit $7$, the purple area of not being digit $7$.}\label{not_circular_dig7_3}
    	\end{minipage}
    	\hfill
    	\begin{minipage}[b]{0.48\textwidth}
    		\includegraphics[width=\textwidth]{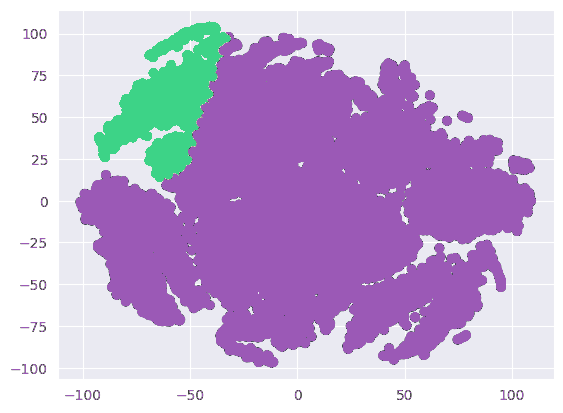}
    		\caption{Binary-classification via {\bf DNN} and $10$ training epochs. The green area corresponds to the digit $7$, the purple area of not being digit $7$.}\label{not_circular_dig7_4}
    	\end{minipage}

    \end{figure}

The parameters (e.g. number of training epochs, learning rate, neural network structure) are kept the same as in the previous experiment (for number $8$). As expected, {\bf ALNN}s (see \autoref{not_circular_dig7_2}) perform similar as {\bf RQNN}s (see \autoref{not_circular_dig7_1}). When using {\bf DNN}s (see \autoref{not_circular_dig7_4}), they work the same as {\bf DRQNN}s (see \autoref{not_circular_dig7_3}).
Therefore, in both favorable and non-favorable cases, using circular layers instead of linear ones doesn't lead to any disadvantages. The full comparison of different types of neural networks can be seen in \autoref{tab:not_circular_dig7_table}.

        \begin{table}
    	\centering
    	\begin{tabular}{c|c|c|c|c|c}
    		Type
    		&
    		Epochs
		&
		Trainable weights
    		&
    		Depth
		&
		Width
    		&
    		Accuracy (Crossentropy)\\
    		\hline
    		{\bf RQNN} & 10 & 3  & 1 & - & 0.9401 \\
    		{\bf DRQNN}  &  10  & 9  & 3 & 5 & 0.94 \\
    		\hline
    		{\bf ALNN} & 10 & 2 & 1 & - & 0.9401 \\
    		{\bf DNN} &   10  & 6 & 3 & 5 & 0.9418 \\
    	\end{tabular}
    	\caption{\label{tab:not_circular_dig7_table} Overview of the result of the numerical experiments for the binary classification of number $7$. One can observe that there is no disadvantage when using circular layers in non-favorable cases. }
    \end{table}

    \section{Conclusion}
    In this work we have show that shallow neural networks with quadratic decision functions are an alternative to deep neural network for binary classification. The algorithms can also be implemented easily based on existing software such as {\bf \emph{Tensorflow}} and {\bf \emph{Keras}} by using customized layers in the implementation. We see an enormous benefit in the problem of detecting subspecies. One should mention that currently different methodologies are used to prove approximation properties of {\bf RQNN}s and {\bf ALNN}s for arbitrary functions. In \cite{FriSchShi24} we used wavelet constructions to prove approximation properties of {\bf RQNN}s, while for {\bf ALNN}s the standard way of proving approximation properties is via the universal approximation theorem (see \cite{Cyb89}). The difference in these two approaches is necessary due to the compactness of the level sets of the according neural network functions.

	\begin{acknowledgement}
This research was funded in whole, or in part, by the Austrian Science Fund
(FWF) 10.55776/P34981 -- New Inverse Problems of Super-Resolved Microscopy (NIPSUM),
and SFB 10.55776/F68  ``Tomography Across the Scales'', project F6807-N36
(Tomography with Uncertainties), and 10.55776/T1160 ``Photoacoustic Tomography: Analysis and Numerics''.
For the purpose of open access, the author has applied a CC BY public copyright
license to any Author Accepted Manuscript version arising from this submission.
The financial support by the Austrian Federal Ministry for Digital and Economic
Affairs, the National Foundation for Research, Technology and Development and the Christian Doppler
Research Association is gratefully acknowledged.

The authors thank Heinz Engl for bringing the referenced newsletter article to their attention.
	\end{acknowledgement}

	
	\section*{References}
	\printbibliography[heading=none]
\end{document}